\documentclass[letterpaper]{article}
\usepackage{aaai25}
\usepackage{times}
\usepackage{helvet}
\usepackage{courier}
\usepackage[hyphens]{url}
\usepackage{graphicx}
\usepackage{natbib}
\usepackage{caption}
\usepackage{amsmath,amssymb,amsfonts}
\usepackage{booktabs}
\usepackage{array}
\usepackage{multirow}
\usepackage[table]{xcolor}
\usepackage{adjustbox}
\usepackage{xspace}
\definecolor{ourrow}{HTML}{EDEDED}

\ifx
\newcommand{\pdfinfo}[1]{}
\fi
\newcommand{\method}{TAPDecoderFT\xspace}
\newcommand{\tap}{TAP\xspace}
\newcommand{\fullft}{FullFT\xspace}
\newcommand{\sourceonly}{SourceOnly\xspace}

\newcommand{\source}{\mathcal{S}}
\newcommand{\target}{\mathcal{T}}
\newcommand{\image}{\mathbf{x}}
\newcommand{\mask}{\mathbf{y}}

\newcommand{\loss}{\mathcal{L}}

\title{Topology-Aware Parameter-Efficient Adaptation\\
for Cross-Dataset Retinal Vessel Segmentation}

\author{
    Yongsong Huang,
    Tomo Miyazaki,
    Kai Xu,
    Xiaofeng Liu\\
    Yaohou Fan,
    Shinichiro Omachi
}
\affiliations{
    Tohoku University,
    Yale University\\
    (hys, tomo, shinichiro.omachi.b5)@tohoku.ac.jp,
    (xu.kai.t6, fan.yaohou.t4)@dc.tohoku.ac.jp,
    xiaofeng.liu@yale.edu
}

\begin{document}
\maketitle

\begin{abstract}
Retinal vessel segmentation in multi-domain deployment requires a source model to adapt to domains that differ in imaging conditions and annotation conventions. Conventional parameter-efficient fine-tuning reduces target-specific storage, but its highly restricted adaptation subspace can be insufficient for reconstructing thin, connected vascular structures. We therefore ask how target-specific capacity should be allocated so that topology-aware supervision remains effective under a
strict per-domain parameter budget. Based on this principle, we propose TAPDecoderFT, a topology-responsive, role-structured adaptation framework. Specifically, TAPDecoderFT shares a fixed source parameter state across deployment domains, uses low-rank residuals for target-specific private/fusion feature mixing, and retains a trainable dense-reconstruction path comprising
the decoder, output head, and refinement module. To promote structurally faithful predictions, the compact target state is jointly optimized with a region-overlap and topology-aware objective that
encourages centerline continuity and thin-branch recovery. It improves both DSC and clDice over GenericLoRA-r4 and narrow TAP-r4 in all six directions and is comparable to full fine-tuning.
\end{abstract}

\section{Introduction}
\label{sec:introduction}

\begin{figure}[!t]
    \centering
    \includegraphics[width=\columnwidth]{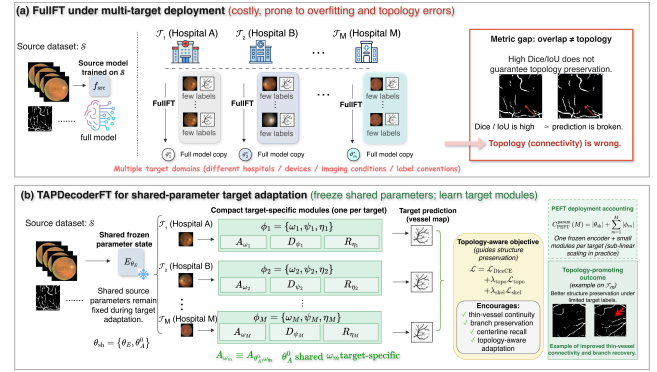}
\caption{Multi-target deployment and shared-state adaptation.
\textbf{(a)} \fullft maintains a complete parameter vector
$\vartheta_m^\star$ for each target; the inset contrasts region overlap with
thin-vessel connectivity. \textbf{(b)} \method reuses a fixed shared state
$\theta_{\mathrm{sh}}$ and learns a compact state $\phi_m$ for each
$\mathcal{T}_m$ under region-overlap and topology-sensitive supervision.}
    \label{fig:intro_challenge_solution}
\end{figure}

Retinal vessel segmentation underpins computational ophthalmology, and
practical deployment typically extends beyond a single curated dataset. A model
trained on one cohort may be transferred to hospitals with different fundus
cameras, illumination profiles, field-of-view conventions, disease
distributions, and annotation habits. Here the \emph{unit of deployment}
becomes the design question. Full fine-tuning (\fullft) adapts a model well to
one target, but serving $M$ domains requires storing and governing $M$ complete
model copies, each with its own training, validation, versioning, and
quality-control cost. When target labels are limited, updating every weight can
also perturb source-trained vascular representations encoding continuity,
caliber variation, and branching geometry. This motivates a different
formulation: \emph{can a fixed source parameter state be shared across
deployment domains and adapted through compact target-specific modules?}

Retinal vessel geometry makes this deployment question especially delicate.
Vessels form thin branching trees with bifurcations, crossings, and terminal
segments; a few missed pixels can sever a centerline, and a few false positives
near the optic disc can hallucinate a branch. Past work has addressed this
structure through matched-filter and ridge-based vessel evidence
\cite{hoover_locating_2000,staal_ridge-based_2004}, modern encoder-decoder
models such as U-Net, Attention U-Net, UNet++, and FR-UNet
\cite{ronneberger_u-net_2015,oktay_attention_2018,zhou_unet_2018,liu_full-resolution_2022},
and recent OCTA or SAM-style medical segmenters with locality-sensitive
enhancement, prompts, or vessel-specific priors
\cite{huang_gaze_2026,chen_sam-octa2_2024,fu_vessam_2025,
ma_segment_2023,zhu_medical_2024}. Although these methods
improve single-domain dense prediction, they do not specify how a source-trained
vessel model should be reused across many small target domains without
maintaining one full model per domain. They also leave a persistent evaluation
gap: \emph{high region overlap does not imply preserved topology}. Dice and IoU measure the pixel-wise overlap between a predicted vessel mask and its annotation, but do not directly test whether the vascular tree remains connected. Because thick vessels account for most foreground pixels, these scores can remain high despite broken centerlines, missing thin branches, or fragmented bifurcations that alter vascular topology. Figure~\ref{fig:intro_challenge_solution} summarizes this shared-state,
topology-aware adaptation setting.

Existing PEFT methods typically choose the trainable parameter subspace
independently of the structural errors that matter for the downstream task.
This separation is problematic for tubular segmentation. Highly restricted
low-rank updates can adapt cross-domain feature mixing, but may expose too
little dense-reconstruction capacity for structural gradients to repair broken
centerlines and missing terminal branches effectively. Consequently, the key
design question is not simply whether to add a topology loss to PEFT, but
\emph{which parameters should remain adaptable so that structural supervision
can act effectively under a constrained per-domain parameter budget}. Prompt
tuning and LoRA-style methods reduce target-specific parameters in medical
segmentation
\cite{fischer_prompt_2022,mandal_sam2lora_2025}, whereas clDice and
skeleton-recall objectives explicitly supervise tubular connectivity
\cite{shit_cldice_2020,kirchhoff_skeleton_2024}. TopoLoRA-SAM combines these
directions for a promptable foundation segmenter
\cite{khazem_topolora-sam_2026}, but does not explicitly study how adaptation
capacity should be allocated between cross-domain feature mixing and dense
vascular reconstruction.

To address these coupled requirements, we propose \method
(Topology-Aware PEFT with a trainable Decoder), which assigns
target-specific capacity according to module role. The encoder parameters and
pretrained private/fusion base kernels remain fixed, low-rank residuals adapt
target-specific feature mixing, and the decoder upsampling blocks, output head,
and refinement path remain trainable for dense reconstruction. Under
parameter-only accounting, the shared state $\theta_{\mathrm{sh}}$ is stored
once, and each deployment domain adds a compact target state $\phi_m$.

This design follows a topology-responsive capacity-allocation principle.
Although compact feature remapping can accommodate cross-domain appearance
changes, correcting missing branches and disconnected centerlines requires
sufficient spatial freedom in the reconstruction pathway. A narrowly adapted
private/fusion endpoint is highly compact ($\sim$0.24\% trainable), but exposes
limited dense-reconstruction capacity. \method therefore optimizes
topology-sensitive supervision within a target-specific subspace that keeps
feature adaptation low-rank while allowing the reconstruction modules to
respond directly to centerline and thin-branch errors.

Together, these components establish a role-aware adaptation paradigm that
preserves a shared source parameter state while allocating target-specific
capacity to cross-domain feature mixing, dense reconstruction, and vascular
structure modeling. Our main contributions are:

$\bullet$ We formulate cross-dataset retinal vessel segmentation as
    structurally supervised optimization over a fixed target-specific
    adaptation subspace, connecting the choice of trainable parameters with the
    topology-sensitive objective.

$\bullet$ We propose \method, a role-structured PEFT framework in which
    low-rank private/fusion residuals handle cross-domain feature mixing, while
    trainable decoder, output-head, and refinement parameters provide the dense
    reconstruction capacity required for centerline and thin-branch correction.

$\bullet$ Across six directed transfers, \method improves both DSC and clDice
    over GenericLoRA-r4 and narrow \tap-r4 in every direction, approaches
    \fullft overlap performance with 18.63\% trainable parameters, and achieves
    higher clDice than standard \fullft in five directions.

\begin{figure}[!t]
    \centering
    \includegraphics[width=\columnwidth]{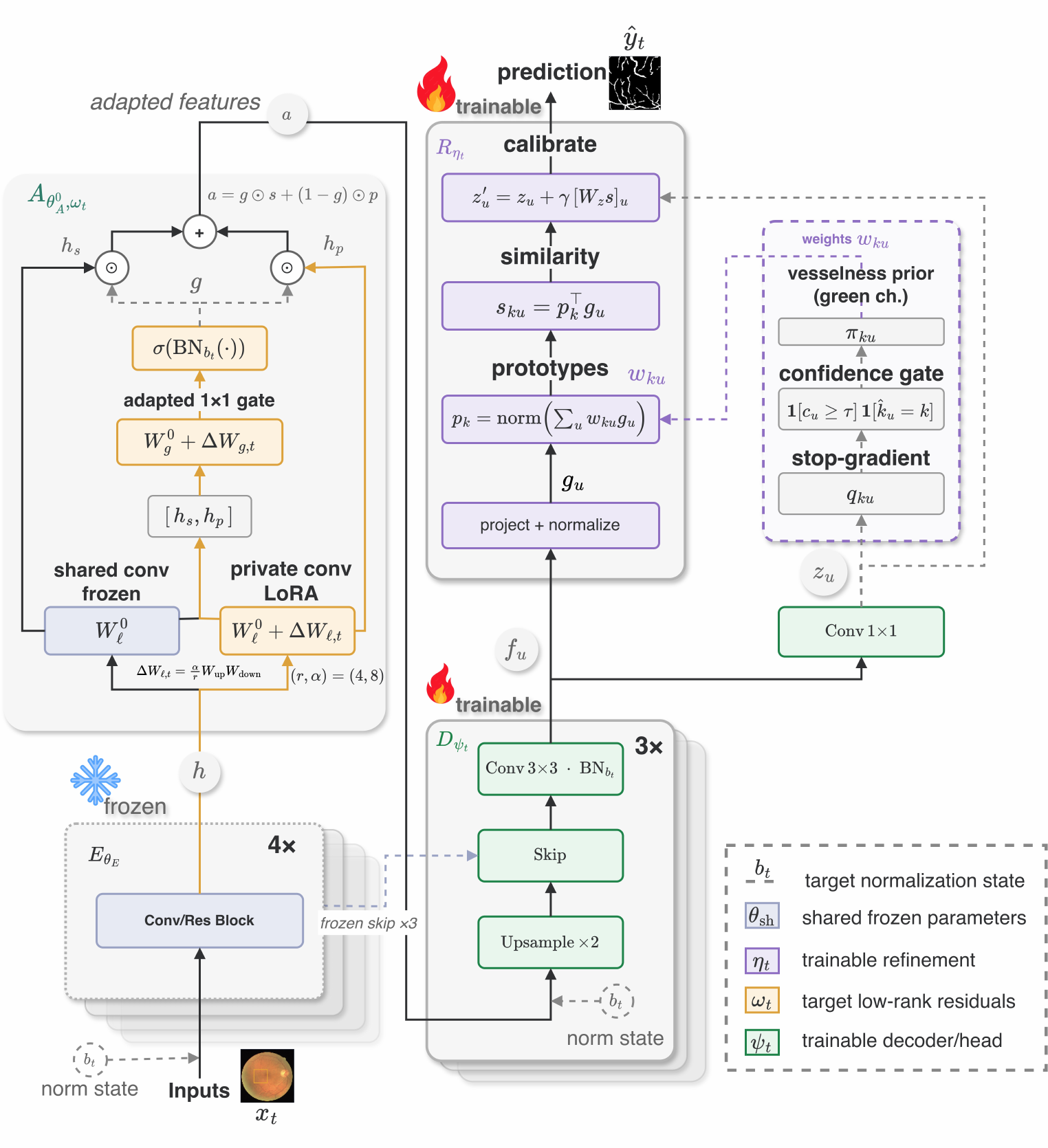}\vspace{-5pt}
\caption{Layer-level parameterization of \method for target domain $t$.
Colors distinguish the frozen shared state
$\theta_{\mathrm{sh}}=\{\theta_E,\theta_A^0\}$ from the trainable target state
$\phi_t=\{\omega_t,\psi_t,\eta_t\}$; dashed links denote the target-specific
normalization state $b_t$. The flow proceeds through low-rank private/fusion
adaptation, decoder reconstruction, and prototype refinement. Only $\phi_t$ is
optimized, while $b_t$ is excluded from parameter accounting.}\vspace{-5pt}
    \label{fig:method_overview}
\end{figure}

\section{Method}
\label{sec:method}

\subsection{Problem Formulation}
Figure~\ref{fig:method_overview} organizes target adaptation into low-rank
private/fusion mixing, decoder reconstruction through frozen encoder skips, and
confidence- and prototype-guided logit calibration. It also separates the
frozen shared state, trainable target state, and non-parameter normalization
state used by these stages.
Let $\source$ be a labeled source retinal vessel dataset and
$\{\target_m\}_{m=1}^{M}$ be the target domains considered at deployment. For a
single target domain, we write $\target$ when the discussion is generic and use
subscript $t$ for the target currently being adapted in a directed transfer.
Each sample is an image-mask pair $(\image,\mask)$, where
$\mask\in\{0,1\}^{H\times W}$ is a binary vessel mask. We partition an adapted model into a frozen parameter state
$\theta_{\mathrm{sh}}=\{\theta_E,\theta_A^0\}$ shared by all targets and a
trainable target parameter state
$\phi_m=\{\omega_m,\psi_m,\eta_m\}$. Here, $\theta_E$ contains the encoder
parameters and $\theta_A^0$ the pretrained base kernels in the private/fusion
paths; $\omega_m$ contains their low-rank residual branches, $\psi_m$ the
decoder and output-head parameters, and $\eta_m$ the prototype-refinement
parameters. We separately denote the target-specific normalization buffers by
$b_m$; these buffers are updated during adaptation but are not included in
$\phi_m$ or in parameter-count accounting. The complete instantiated model has
$P=|\theta_{\mathrm{sh}}|+|\phi_m|$ scalar parameters. Throughout the parameter
formulas, $|\cdot|$ and
$C^{\mathrm{param}}$ count model parameters only; optimizer state,
normalization running statistics, and other buffers are excluded. A source model is first trained on $\source$. For a target training set
$\target_{\mathrm{tr}}=\{(\image_i,\mask_i)\}_{i=1}^{N_t}$, standard full
fine-tuning treats the complete parameter vector
$\vartheta\in\mathbb{R}^{P}$ as target-specific and solves
\begin{equation}
    \vartheta_t^\star
    =
    \arg\min_{\vartheta\in\mathbb{R}^{P}}
    \frac{1}{N_t}\sum_{i=1}^{N_t}
    \mathcal{L}_{\mathrm{DiceCE}}
    \big(f_{\vartheta}(\image_i),\mask_i\big).
\end{equation}

In contrast, \method restricts the target-specific parameters to the
role-structured adaptation subspace
\begin{equation}
    \Phi_{\mathrm{TAPDec}}
    =
    \Omega_r \times \Psi_D \times \mathcal{H}_R ,
\end{equation}
where $\Omega_r$ contains the rank-$r$ residual parameters in the private and
fusion pathways, $\Psi_D$ contains the decoder and output-head parameters, and
$\mathcal{H}_R$ contains the prototype-refinement parameters. Its trainable
dimension is
\begin{equation}
    d_{\mathrm{TAPDec}}
    =
    |\omega_t|+|\psi_t|+|\eta_t|,
    \qquad
    \rho=\frac{d_{\mathrm{TAPDec}}}{P}.
\end{equation}

Target adaptation optimizes the structural segmentation objective within this
fixed parameter subspace:
\begin{equation}
    \begin{aligned}
    \phi_t^\star
    &=
    \arg\min_{\phi_t\in\Phi_{\mathrm{TAPDec}}}
    \frac{1}{N_t}\sum_{i=1}^{N_t} \\
    &\quad
    \mathcal{L}_{\mathrm{TAP}}
    \left(
    f_{\theta_{\mathrm{sh}},\phi_t}(\image_i;b_t),\mask_i
    \right).
    \end{aligned}
\end{equation}
Here $\phi_t$ is the gradient-optimized parameter state, whereas $b_t$ denotes
the target-specific normalization buffers updated by the corresponding
normalization rules. The buffers are not model parameters and are excluded
from $\Phi_{\mathrm{TAPDec}}$, $d_{\mathrm{TAPDec}}$, and all parameter-count
comparisons.
For rank-4 \method, $P=13{,}734{,}921$,
$d_{\mathrm{TAPDec}}=2{,}559{,}017$, and
$|\theta_{\mathrm{sh}}|=11{,}175{,}904$, giving
$\rho=0.186315$.
For $M$
equal-size target modules, parameter-copy accounting is
\begin{equation}
    \begin{aligned}
    C_{\mathrm{FullFT}}^{\mathrm{param}}(M)&=MP,\\
    C_{\mathrm{PEFT}}^{\mathrm{param}}(M)&=|\theta_{\mathrm{sh}}|+
    \sum_{m=1}^{M}|\phi_m|.
    \end{aligned}
\end{equation}
Using $|\phi_m|=\rho P$ and
$|\theta_{\mathrm{sh}}|=(1-\rho)P$ yields
\begin{equation}
 \begin{aligned}
 \frac{C_{\mathrm{PEFT}}^{\mathrm{param}}(M)}
 {C_{\mathrm{FullFT}}^{\mathrm{param}}(M)}
 &=\rho+\frac{1-\rho}{M},\\
 \mathrm{Saving}(M)&=(1-\rho)\!\left(1-\frac{1}{M}\right).
 \end{aligned}
\end{equation}
As an illustrative multi-target deployment scenario, when $M=10$, the resulting storage ratio is 26.77\%, a 73.23\%
parameter-count reduction. This calculation uses a common instantiated size
$P$; the measured \fullft implementation omits 31,360 adapter parameters, and
using its exact denominator changes the ratio only to 26.83\% (73.17\%
reduction). The accounting does not
measure storage systems or regulatory cost directly. The current implementation
updates normalization running statistics during adaptation and stores them as a
separate target-specific non-parameter state $b_m$. These buffers are excluded
from $C^{\mathrm{param}}$, as are all other non-parameter states. All model buffers together contain
11,873 scalars (approximately 47.5~kB in FP32, or 0.087\% as many scalars as
model parameters), so they do not change the deployment scaling argument.

At inference time, $\theta_{\mathrm{sh}}$ is stored once, and $(\phi_m,b_m)$ is
selected according to the deployment domain. For target $\mathcal{T}_m$,
prediction uses $f_m(x;b_m)=f_{\theta_{\mathrm{sh}},\phi_m}(x;b_m)$.
Target modules do not interact during inference; adding a new deployment domain
therefore adds a compact $\phi_m$ and small buffer state rather than another
complete model copy.

\subsection{Source Model and Adapter Placement}
We instantiate the source model using the DCD-Retina dual-branch
encoder--decoder architecture \cite{cong_divide-and-conquer_2026}, which
contains shared and private feature pathways, multi-scale fusion gates, a
bottleneck, decoder upsampling blocks, an output head, and prototype
refinement. \method adapts this architecture by separating its parameters into
a shared source state and a compact target-specific state, as illustrated in
Figure~\ref{fig:method_overview}.  Using the shared boundary defined above, \method decomposes the model as
\begin{equation}
f_{\theta_{\mathrm{sh}},\phi_t}(\cdot;b_t)
=
R_{\eta_t}\circ
D_{\psi_t}\circ
A_{\theta_A^0,\omega_t,b_t}\circ
E_{\theta_E,b_t}(\cdot),
\end{equation}
and trains only the target parameter state
$\phi_t=\{\omega_t,\psi_t,\eta_t\}$.

For an input feature map $h$, each adapted private/fusion layer is implemented
as
\begin{equation}
\operatorname{Conv}_{\ell,t}(h)
=
W_\ell^0*h
+
\frac{\alpha}{r}
W_{\mathrm{up},\ell,t}*
\left(
W_{\mathrm{down},\ell,t}*h
\right),
\end{equation}
where the base convolution $W_\ell^0$ is frozen and
$W_{\mathrm{down},\ell,t}$ and $W_{\mathrm{up},\ell,t}$ are target-specific
$1\times1$ projections. The residual branch therefore has channel-mixing rank
at most $r$ while leaving the spatial coefficients of the base convolution
unchanged.

For a fusion gate, we set $h=[h_s,h_p]$ and compute
\begin{equation}
\begin{aligned}
g &=
\sigma\!\left(
\mathrm{BN}_{b_t}
\left(
\operatorname{Conv}_{g,t}([h_s,h_p])
\right)
\right), \\
a &= g\odot h_s+(1-g)\odot h_p.
\end{aligned}
\end{equation}

We inject these residual branches into
\texttt{private\_downs}, \texttt{private\_bottleneck},
\texttt{fusion\_gates}, and \texttt{bottleneck\_fusion}. This role-structured
partition assigns target-specific feature mixing to the low-rank residuals and
dense reconstruction to the decoder, output head, and refinement module. We use
$(r,\alpha)=(4,8)$.

\subsection{Confidence- and Prototype-Guided Refinement}
Let $z_u\in\mathbb{R}^{2}$ be the coarse logits and $f_u$ the decoder feature
at pixel $u$. The refinement module projects and normalizes the feature as
$g_u=\operatorname{norm}(W_f f_u)$ and obtains detached class probabilities
$q_{ku}=\operatorname{softmax}(z_u)_k$. A green-channel vesselness prior is
computed by
\begin{equation}
 v_u=\operatorname{minmax}\!\left(
 [\operatorname{AvgPool}_{15}(x^G)_u-x^G_u]_+\right),
\end{equation}
with class priors $\pi_{1u}=1+\beta v_u$ and
$\pi_{0u}=1+\beta(1-v_u)$. Writing
$c_u=\max_k q_{ku}$ and $\hat{k}_u=\arg\max_k q_{ku}$, the confident prototype
weights are
\begin{equation}
 \begin{aligned}
 \widetilde w_{ku}&=q_{ku}\,\mathbf{1}[c_u\ge\tau]\,
 \mathbf{1}[\hat{k}_u=k]\,\pi_{ku},\\
 w_{ku}&=\widetilde w_{ku}/
 (\sum_v\widetilde w_{kv}+\epsilon).
 \end{aligned}
\end{equation}
The class prototype and calibrated logits are then
\begin{equation}
 \begin{aligned}
 p_k&=\operatorname{norm}\!\left(\sum_u w_{ku}g_u\right),
 &s_{ku}&=p_k^\top g_u,\\
 z'_u&=z_u+\gamma[W_zs]_u.&&
 \end{aligned}
\end{equation}
We use $\tau=0.7$, $\beta=0.5$, and a trainable $\gamma$ initialized to 0.05.
When a class contains too few confident pixels, the implementation falls back
to probability-weighted prototypes. Detaching $q$ prevents gradients through prototype
selection, while $W_f$, $W_z$, and $\gamma$ remain trainable.

\subsection{Topology-Responsive Adaptation Subspace}

\paragraph{Low-rank feature mixing.}
The narrow \tap endpoint restricts low-rank adaptation to the private and
fusion pathways while keeping the decoder upsampling blocks fixed; only the
output head and prototype-refinement module remain trainable. This endpoint
updates 0.2363\% of the parameters but reaches a mean DSC of 0.7197, indicating
that highly restricted adaptation provides insufficient capacity for dense
cross-dataset reconstruction. \method therefore retains the low-rank
private/fusion residuals while additionally training the decoder upsampling
blocks, output head, and prototype refinement. Unlike GenericLoRA's broad
low-rank placement, this role-structured design assigns target-specific feature
mixing to restricted residuals and dense reconstruction to the decoder path.

\paragraph{Topology-responsive dense reconstruction.}
The update sets are not strictly nested, so we compare their trainable
dimensions rather than assert set inclusion. For
$d_q=|\mathcal{I}_q|$, the implemented variants satisfy
\begin{equation}
 \begin{aligned}
 d_{\mathrm{head}}
 &< d_{\mathrm{nTAP}}
 < d_{\mathrm{dec}}
 = d_{\mathrm{dec+topo}}\\
 &< d_{\mathrm{dec+adp}}
 = d_{\mathrm{TAPDec}}
 < d_{\mathrm{FullFT}}
 = P.
 \end{aligned}
\end{equation}
Here ``topo'' changes only the training objective, whereas ``adp'' adds the
private/fusion low-rank branches. This ordering places \method between narrow
adaptation and \fullft while exposing the decoder as the principal
target-specific dense-reconstruction pathway.

\subsection{Topology-Aware Objective}

The optimization in the preceding formulation uses a joint region- and
structure-sensitive objective. For clarity, we next define its topology terms.
The shared state $\theta_{\mathrm{sh}}$ remains fixed, and all gradients act
through the compact target state
$\phi_t=\{\omega_t,\psi_t,\eta_t\}$.

The adaptation loss combines a standard region-overlap segmentation objective
with topology-sensitive regularization:
\begin{equation}
\mathcal{L}_{\mathrm{TAP}}
=
\mathcal{L}_{\mathrm{DiceCE}}
+
\lambda_{\mathrm{topo}}\mathcal{L}_{\mathrm{topo}}
+
\lambda_{\mathrm{skel}}\mathcal{L}_{\mathrm{skel}}.
\end{equation}
The selected setting uses $\lambda_{\mathrm{topo}}=0.1$ and
$\lambda_{\mathrm{skel}}=0.05$. Let
$p=f_{\theta_{\mathrm{sh}},\phi_t}(\image;b_t)$ be the soft vessel
probability map, and let $S(\cdot)$ denote a differentiable soft-skeleton
operator. Our implementation uses 20 iterations of soft morphological
erosion and opening.
Following the clDice formulation
\cite{shit_cldice_2020}, we define topological precision and sensitivity as
\begin{equation}
    \begin{aligned}
    T_{\mathrm{prec}}(p,\mask)
    &=
    \frac{\langle S(p),\mask\rangle+\epsilon}
         {\|S(p)\|_1+\epsilon}, \\
    T_{\mathrm{sens}}(p,\mask)
    &=
    \frac{\langle S(\mask),p\rangle+\epsilon}
         {\|S(\mask)\|_1+\epsilon}.
    \end{aligned}
\end{equation}
The corresponding clDice score is
\begin{equation}
    \mathrm{clDice}(p,\mask)
    =
    \frac{2T_{\mathrm{prec}}(p,\mask)T_{\mathrm{sens}}(p,\mask)}
    {T_{\mathrm{prec}}(p,\mask)+T_{\mathrm{sens}}(p,\mask)+\epsilon}.
\end{equation}

We define
$\loss_{\mathrm{topo}}=1-\mathrm{clDice}(p,\mask)$ and
$\loss_{\mathrm{skel}}=1-T_{\mathrm{sens}}(p,\mask)$. The clDice term balances
centerline precision and sensitivity, penalizing both spurious and missing
vascular skeletons. The additional skeleton-recall term places greater emphasis
on recovering the reference centerline, whose thin and terminal segments are
particularly vulnerable under cross-dataset adaptation. Together, these terms
make vascular structure an explicit optimization target rather than a
post-hoc evaluation criterion.

\section{Experiments}
\label{sec:experiments}

\subsection{Datasets and Transfer Protocol}
We evaluate binary retinal vessel segmentation on DRIVE, CHASE\_DB1, and STARE.
The transfer matrix contains six directed settings:
DRIVE$\rightarrow$CHASE\_DB1, CHASE\_DB1$\rightarrow$DRIVE,
DRIVE$\rightarrow$STARE, STARE$\rightarrow$DRIVE,
CHASE\_DB1$\rightarrow$STARE, and STARE$\rightarrow$CHASE\_DB1.
Three-class RAVIR artery/vein segmentation is outside the main claim of this
paper.

\paragraph{Experimental settings.}
Table~\ref{tab:optimization_protocol} summarizes the stage-specific settings.
All transfer methods share the 20-epoch budget, augmentation, StepLR form, and
checkpoint rule.
Training and validation curves reach a stable plateau within this budget. For STARE, we use nested leave-one-out evaluation. In each outer fold, one image
is reserved exclusively for testing, one of the remaining 19 images is used
for validation, and the other 18 images are used for training or adaptation.

\begin{table}[t]
\centering
\caption{Training settings. All runs use AdamW, batch size 2, and
validation-best DSC checkpoints. Ep./seed and StepLR denote epochs/seed and
step size/factor.}\vspace{-5pt}
\label{tab:optimization_protocol}
\scriptsize
\setlength{\tabcolsep}{2.1pt}
\begin{adjustbox}{max width=\columnwidth}
\begin{tabular}{@{}lcccc@{}}
\toprule
Stage & Ep./seed & Initial LR & WD & StepLR \\
\midrule
Source pretraining & 50/42 & $8\times10^{-4}$ & $10^{-2}$ & 25/0.5 \\
Direct DCD & 50/42 & $8\times10^{-4}$ & $10^{-2}$ & 25/0.5 \\
Direct Attn. U-Net & 50/42 & $10^{-3}$ & $10^{-2}$ & 25/0.5 \\
Adapt. \fullft & 20/1 & $5\times10^{-5}$ & $10^{-4}$ & 5/0.5 \\
Adapt. LoRA/\tap & 20/1 & $10^{-4}$ & $10^{-4}$ & 5/0.5 \\
Adapt. \method & 20/1 & $3\times10^{-4}$ & $10^{-4}$ & 5/0.5 \\
\bottomrule\vspace{-5pt}
\end{tabular}
\end{adjustbox}
\end{table}

Method-family learning rates were fixed across directions: $5\times10^{-5}$ for
\fullft and $10^{-4}$ for GenericLoRA/narrow \tap. For \method, 20-epoch
development runs on DRIVE$\rightarrow$CHASE\_DB1 compared
$\{5\times10^{-5},10^{-4},2\times10^{-4},3\times10^{-4}\}$; $3\times10^{-4}$
gave the highest validation DSC and was then fixed for all directions. No
direction-specific retuning was performed. Because this direction is also
reported, the selection is not fully held out.

\subsection{Baselines and Metrics}
The main comparison focuses on recent or directly relevant adaptation methods:
GenericLoRA-r4\cite{hu_lora_2022} as a general low-rank PEFT baseline. Narrow \tap-r4 is our controlled
role-restricted adapter baseline; it uses the same low-rank parameterization
with topology and skeleton-recall terms based on
\cite{shit_cldice_2020,kirchhoff_skeleton_2024}. Source-only inference and \fullft are retained as lower and upper
deployment anchors. We additionally report directly target-trained Attention U-Net
\cite{oktay_attention_2018} and DCD-Retina
\cite{cong_divide-and-conquer_2026} variants, evaluated using the same
region-overlap and topology-sensitive metrics. The strongest such baselines per target are summarized in
Table~\ref{tab:direct_train_reference}. Direct training is not the central paired
baseline for the deployment claim, since it maintains a separate model per target
rather than adapting a shared source model.

We report DSC and IoU for region overlap, together with three complementary
structure-sensitive measures. clDice quantifies agreement between predicted and
reference centerlines; skeleton recall measures recovery of the reference
vascular skeleton; and thin-vessel Dice evaluates segmentation within
low-caliber regions identified by iterative erosion. We additionally report
the discrepancy between predicted and reference 8-connected component counts.
These measures evaluate distinct aspects of structural fidelity that are not
captured by region overlap alone.

The direct-train reference table does not
include thin-vessel Dice, so thin-vessel direct-train deltas are not reported.
Error bars in the summary figures describe variation across transfer directions,
not repeated-run uncertainty.

\begin{table}[t]
\centering
\caption{Mean results across 6 directed binary transfers. DecoderFT-only
and DecoderFT+Topo isolate the topology supervision at the
same trainable-parameter budget.}\vspace{-5pt}
\label{tab:main_results}
\scriptsize
\setlength{\tabcolsep}{2.2pt}
\begin{adjustbox}{max width=\columnwidth}
\begin{tabular}{@{}lrrrrr@{}}
\toprule
Method & Train. \% & DSC & IoU & clDice & Thin Dice \\
\midrule
\sourceonly & 0.0000 & 0.4366 & 0.3059 & 0.4008 & 0.4129 \\
GenericLoRA-r4 & 0.4829 & 0.7248 & 0.5712 & 0.7050 & 0.7772 \\
\tap-r4 & 0.2363 & 0.7197 & 0.5649 & 0.6987 & 0.7716 \\
\midrule
DecoderFT-only & 18.4453 & 0.7685 & 0.6249 & 0.7673 & 0.8268 \\
DecoderFT+Topo & 18.4453 & 0.7702 & 0.6272 & 0.7778 & 0.8414 \\
\rowcolor{ourrow}
\method-r4 & 18.6315 & 0.7722 & 0.6297 & \textbf{0.7809} & 0.8444 \\
\midrule
\fullft & 100.0000 & \textbf{0.7800} & \textbf{0.6398} & 0.7791 & \textbf{0.8471} \\
\bottomrule
\end{tabular}\vspace{-5pt}
\end{adjustbox}
\end{table}

\begin{table*}[t]
\centering
\caption{Selected source-target adaptation matrix for recent PEFT-style
baselines. Each cell reports DSC/clDice for one directed transfer. The table
shows that \method consistently improves over GenericLoRA-r4 and narrow
\tap-r4, while the \fullft comparison remains direction-dependent.}
\label{tab:selected_direction_matrix}
\begin{adjustbox}{max width=\textwidth}
\begin{tabular}{lrrrrrr>{\columncolor{ourrow}}r>{\columncolor{ourrow}}r}
\toprule
 & \multicolumn{2}{c}{\fullft} & \multicolumn{2}{c}{GenericLoRA-r4} &
 \multicolumn{2}{c}{\tap-r4} & \multicolumn{2}{c}{\cellcolor{ourrow}\method-r4} \\
\cmidrule(lr){2-3}\cmidrule(lr){4-5}\cmidrule(lr){6-7}\cmidrule(lr){8-9}
Direction & DSC & clDice & DSC & clDice & DSC & clDice & DSC & clDice \\
\midrule
DRIVE$\rightarrow$CHASE\_DB1 & 0.7733 & 0.7735 & 0.7219 & 0.7198 & 0.7165 & 0.7172 & \textbf{0.7749} & \textbf{0.7834} \\
CHASE\_DB1$\rightarrow$DRIVE & \textbf{0.7828} & 0.7622 & 0.7420 & 0.7145 & 0.7323 & 0.7041 & 0.7819 & \textbf{0.7722} \\
DRIVE$\rightarrow$STARE & \textbf{0.7697} & \textbf{0.7851} & 0.6236 & 0.5801 & 0.6213 & 0.5738 & 0.7332 & 0.7510 \\
STARE$\rightarrow$DRIVE & \textbf{0.7929} & 0.7778 & 0.7721 & 0.7444 & 0.7676 & 0.7373 & 0.7906 & \textbf{0.7861} \\
CHASE\_DB1$\rightarrow$STARE & \textbf{0.7792} & 0.7942 & 0.7429 & 0.7353 & 0.7365 & 0.7257 & 0.7709 & \textbf{0.8044} \\
STARE$\rightarrow$CHASE\_DB1 & \textbf{0.7819} & 0.7819 & 0.7464 & 0.7359 & 0.7442 & 0.7344 & 0.7818 & \textbf{0.7884} \\
\bottomrule
\end{tabular}
\end{adjustbox}
\end{table*}

\subsection{Main Transfer Results}
Tables~\ref{tab:main_results} and~\ref{tab:selected_direction_matrix} summarize
the primary accuracy results; Source-only inference collapses across domains
(mean DSC 0.4366), confirming substantial cross-dataset shift. GenericLoRA-r4
and narrow \tap-r4 remain below \fullft on dense reconstruction. In every
direction, \method improves both DSC and clDice over these two PEFT baselines,
raising their mean DSC by 0.0474/0.0525 and mean clDice by 0.0759/0.0822,
respectively.

The comparison with \fullft reveals a clear parameter--accuracy trade-off.
\method is lower by 0.0078 mean DSC and 0.0101 mean IoU, but recovers most of
the dense-prediction capability of \fullft while updating only 18.63\% of the
parameters. On topology-sensitive evaluation, \method reaches a mean clDice of
0.7809 versus 0.7791 for \fullft and achieves higher clDice in five of six
directions. Overall, these results reveal a favorable adaptation regime in which \method
retains near-\fullft region overlap with substantially fewer target-specific
parameters while providing stronger topology-sensitive behavior across most
transfer directions.

The controlled comparison further isolates the source of this structural gain.
At the same 18.4453\% trainable-parameter budget, adding topology supervision
to DecoderFT increases mean clDice from 0.7673 to 0.7778 and thin-vessel Dice
from 0.8268 to 0.8414. Adding the restricted private/fusion adapters yields the
highest overall clDice of 0.7809. Together, these results show that decoder
adaptation restores dense reconstruction capacity, while topology supervision
provides the primary improvement in centerline-sensitive behavior.

Across the four transfers with DRIVE or CHASE\_DB1 as the target,
\method exceeds \fullft at 10/25/50\% target labels by
$+0.0046/+0.0010/+0.0013$ DSC and $+0.0162/+0.0108/+0.0144$ clDice. At 100\%,
DSC is 0.7823 versus 0.7827, whereas clDice remains higher (0.7825 versus
0.7739).

\begin{table*}[t]
\centering
\caption{Target-trained references and the highest-DSC observed incoming
\method transfer for each target. For each target, the displayed transfer is
selected retrospectively from the two available source domains using target-test
DSC and is reported only as descriptive context rather than as the primary
paired adaptation benchmark.}
\label{tab:direct_train_reference}
\begin{adjustbox}{max width=\textwidth}
\begin{tabular}{lllrrrrr}
\toprule
Target & Setting & Model or source & DSC & IoU & clDice & Skel. rec. & Comp. err. \\
\midrule
DRIVE & direct train & Attention U-Net & 0.7899 & 0.6528 & 0.7687 & 0.4923 & 30.6500 \\
DRIVE & direct train & DCD-Retina-CP & 0.7896 & 0.6523 & 0.7616 & 0.4838 & 35.1000 \\
DRIVE & direct train & DCD-Retina-NP & 0.7819 & 0.6419 & 0.7573 & 0.4844 & 25.1000 \\
\rowcolor{ourrow} DRIVE & \method transfer & STARE$\rightarrow$DRIVE & 0.7906 & 0.6540 & 0.7861 & 0.4291 & 42.5000 \\
\midrule
CHASE\_DB1 & direct train & DCD-Retina w/o Gate & 0.7962 & 0.6614 & 0.7791 & 0.4817 & 31.5000 \\
CHASE\_DB1 & direct train & DCD-Retina-NP & 0.7938 & 0.6580 & 0.7760 & 0.4822 & 35.9286 \\
CHASE\_DB1 & direct train & DCD-Retina-CP & 0.7930 & 0.6570 & 0.7720 & 0.4761 & 31.8571 \\
\rowcolor{ourrow} CHASE\_DB1 & \method transfer & STARE$\rightarrow$CHASE\_DB1 & 0.7818 & 0.6426 & 0.7884 & 0.3993 & 41.9286 \\
\midrule
STARE & direct train & DCD-Retina-CP & 0.8008 & 0.6715 & 0.7989 & 0.5509 & 45.3500 \\
STARE & direct train & DCD-Retina w/o Gate & 0.7997 & 0.6705 & 0.7982 & 0.5462 & 48.2000 \\
STARE & direct train & DCD-Retina w/o Private & 0.7992 & 0.6695 & 0.7959 & 0.5461 & 47.8500 \\
\rowcolor{ourrow} STARE & \method transfer & CHASE\_DB1$\rightarrow$STARE & 0.7709 & 0.6272 & 0.8044 & 0.4547 & 59.0000 \\
\bottomrule
\end{tabular}
\end{adjustbox}
\end{table*}

\subsection{Target-Trained Baseline Context}
Table~\ref{tab:direct_train_reference} contextualizes transfer against models
trained directly on each target. The highest-DSC observed incoming \method transfer matches the best
direct DSC on DRIVE but trails by 0.0144/0.0299 on CHASE\_DB1/STARE; conversely,
it has the highest clDice on all three targets. Lower skeleton recall and higher
component error bound this topology interpretation. Because both direct training and transfer are evaluated using the same outer
STARE folds, their target-test metrics are aligned at the dataset level;
however, direct training and source-to-target adaptation address different
deployment settings and are therefore interpreted as complementary baselines.

\subsection{Ablation Studies}

A controlled $2\times2$ comparison independently toggles private/fusion adaptation and topology
supervision. Starting from DecoderFT-only, topology supervision improves mean
clDice by 0.0105 and thin-vessel Dice by 0.0145, demonstrating that the
structural objective produces gains beyond those obtained from reconstruction
capacity alone. The restricted adapters provide complementary improvements in
region overlap and feature adaptation. Combining both components yields the
highest mean DSC, clDice, and thin-vessel Dice
(0.7722/0.7809/0.8444), outperforming DecoderFT-only by
0.0037/0.0136/0.0176. Importantly, the complete model improves all three
metrics over DecoderFT-only in every transfer direction, supporting the
complementarity of role-aware capacity allocation and topology-aware
supervision. Adapters alone improve mean DSC/clDice/thin-vessel Dice by 0.0019/0.0019/0.0024; adding topology supervision yields 0.0018/0.0117/0.0152, with larger gains in structural metrics than overlap.

The same comparison also locates the main capacity source. Relative to narrow
\tap-r4 in Table~\ref{tab:main_results}, DecoderFT-only improves mean DSC from
0.7197 to 0.7685 even before adapters or topology terms are added. The decoder,
output head, and refinement path therefore supply most of the dense
reconstruction capacity, while the two proposed factors refine that operating
point. The rank sweep on
DRIVE$\leftrightarrow$CHASE\_DB1 shows little benefit from increasing rank:
ranks 1 to 8 differ by at most 0.0008 DSC and 0.0018 clDice on average. A parallel
scaling sweep at fixed rank~4 is equally flat: varying $\alpha\in\{4,8,16\}$
(effective scale $\alpha/r\in\{1,2,4\}$) changes mean DSC by at most 0.0017 and
clDice by at most 0.0007. We use $(r,\alpha)=(4,8)$ as a robust default rather
than treating it as a tuned optimum.

\begin{figure}[t]
    \centering
    \includegraphics[width=\columnwidth]
    {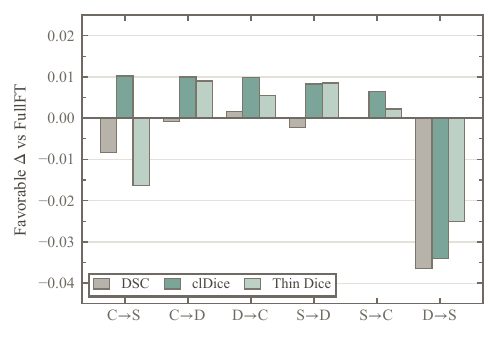}\vspace{-5pt}
    \caption{Direction-level structural trade-off relative to \fullft
    (positive values favor \method). \method achieves higher clDice in five
    transfer directions while maintaining comparable DSC in most settings.
    DRIVE$\rightarrow$STARE (D$\rightarrow$S) exhibits the largest observed
    appearance shift.}
    \label{fig:topology_tradeoff}
\end{figure}

\begin{figure}[t]
    \centering
    \includegraphics[width=\columnwidth]{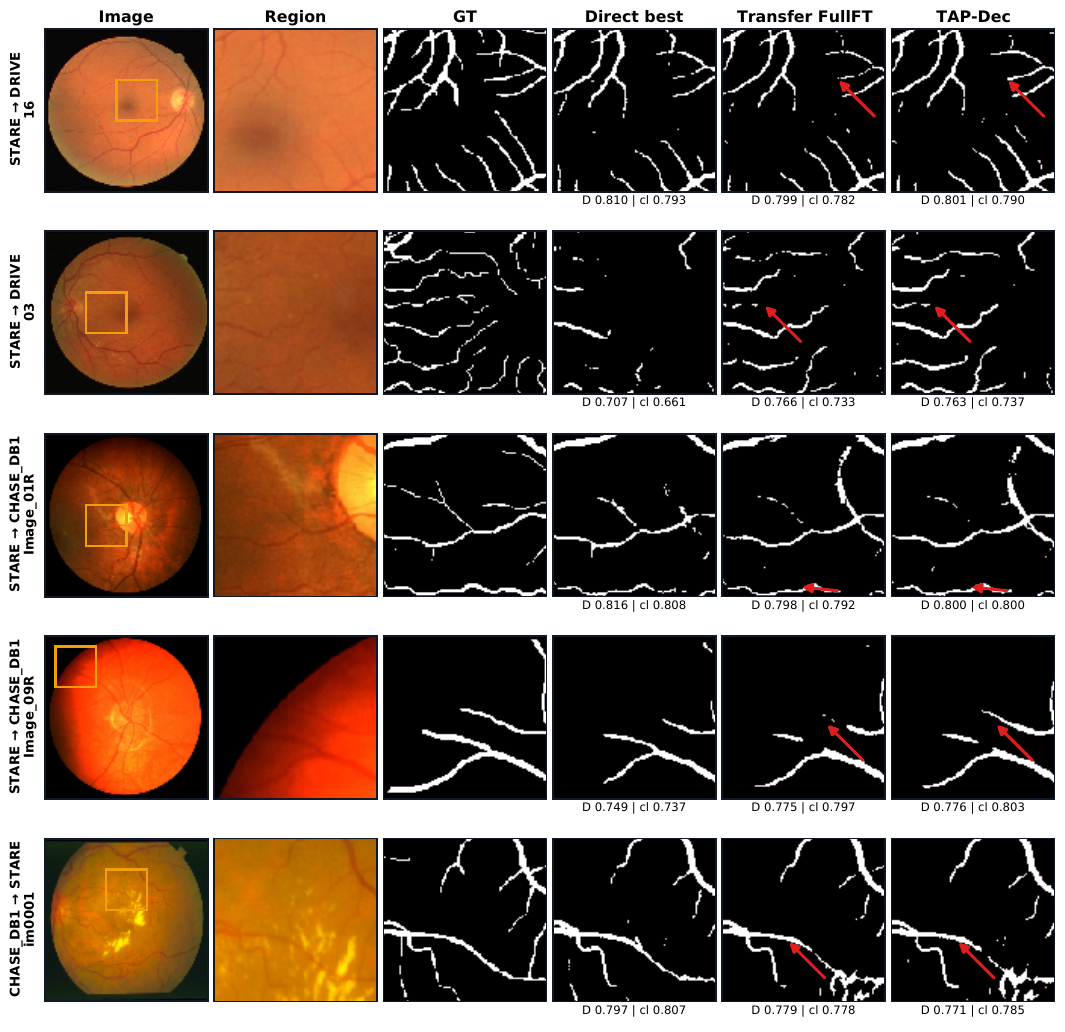}\vspace{-5pt}
    \caption{Qualitative transfer examples. Each row shows a local vessel region
    and masks from the ground truth, direct-trained reference, transfer
    \fullft, and \method; the examples highlight how \method preserves thin-branch continuity and
reduces local fragmentation under cross-dataset adaptation.}
    \label{fig:qualitative}
\end{figure}

\subsection{Topology Trade-off}
Figure~\ref{fig:topology_tradeoff} shows direction-level deltas. In
DRIVE$\rightarrow$ CHASE\_DB1, \method slightly exceeds \fullft on DSC
(0.7749 versus 0.7733), clDice (0.7834 versus 0.7735), and thin-vessel Dice
(0.8561 versus 0.8507). In CHASE\_DB1$\rightarrow$DRIVE, \method loses only
0.0009 DSC while gaining about 0.0100 clDice. In STARE$\rightarrow$DRIVE, it
loses about 0.0023 DSC while gaining about 0.0083 clDice. In
STARE$\rightarrow$CHASE\_DB1, DSC is essentially tied and clDice improves by
about 0.0065.

The largest gap occurs for DRIVE$\rightarrow$STARE, where \method trails
\fullft by 0.0365 DSC. This direction likely requires stronger adaptation of
the frozen spatial filters than the current private/fusion residuals provide.
Nevertheless, \method substantially outperforms both GenericLoRA-r4 and narrow
\tap-r4 in this setting, indicating that decoder-aware capacity allocation
remains beneficial even under the strongest observed shift.

The direction-level consistency of these improvements is reported in
Table~\ref{tab:selected_direction_matrix}.

\subsection{Qualitative Analysis}
Figure~\ref{fig:qualitative} complements the quantitative metrics by showing
local vessel regions where overlap scores can hide topology-sensitive errors. The visual cases correspond to the deployment issues introduced at the beginning
of the paper: cross-domain adaptation often produces broken centerlines, missing
terminal branches, and small fragmented components that are not fully reflected
by DSC or IoU. Comparing masks within each row shows how decoder-aware target
adaptation changes these local structures. In the favorable examples, \method
more closely follows the annotated thin branches and closes local gaps relative
to the transfer alternatives, which visually explains the gains in clDice and
thin-vessel behavior reported in the quantitative tables. In the STARE-target example, however, several low-contrast terminal branches remain absent, consistent with the lower scores for DRIVE$\rightarrow$STARE. These residual errors indicate that decoder-aware adaptation improves branch continuity in favorable transfers but does not fully compensate for substantial target-domain shift.

\section{Conclusion}
\label{sec:conclusion}

In this study, we presented \method, a topology-aware, role-structured parameter-efficient
adaptation strategy for cross-dataset retinal vessel segmentation. By coupling
low-rank private/fusion feature adaptation with trainable dense reconstruction
and explicit centerline supervision, \method addresses domain shift,
target-specific storage, and vascular structural fidelity within a unified
framework. With only 18.63\% trainable parameters, it approaches the mean
region-overlap performance of \fullft, achieves higher clDice in five of six
transfer directions, and consistently outperforms the evaluated low-rank PEFT
baselines. Controlled ablations further show that decoder adaptation restores
dense prediction capacity, while topology supervision provides the principal
gain in centerline-sensitive and thin-vessel metrics. These results establish
role-aware capacity allocation as an effective design principle for
multi-target adaptation of dense tubular segmenters.

\bibliography{references}

\end{document}